\documentclass{article}

\usepackage{graphicx}
 \usepackage[nonatbib, final]{neurips_2021_ml4ps}

\usepackage[utf8]{inputenc} 
\usepackage[T1]{fontenc}    
\usepackage{hyperref}       
\usepackage{url}            
\usepackage{booktabs}       
\usepackage{amsfonts, amsmath}       
\usepackage{nicefrac}       
\usepackage{microtype}      
\usepackage{xcolor} 
\usepackage{adjustbox}

\graphicspath{ {./images/} }

\title{Fine-tuning Vision Transformers for the Prediction of State Variables in Ising Models}

\author{Onur Kara\thanks{Hindsight Technology Solutions}\\
 \texttt{okara83@gmail.com}
    \And    
    Arijit Sehanobish\thanks{Work done as a postdoc at Yale University. Currently at Covera Health, NYC}\\
    \texttt{arijit.sehanobish1@gmail.com}
    \And    
    Hector H. Corzo\thanks{ Center for Chemical Computation and Theory, University of California, Merced, CA.}\\
    \texttt{hhcorzo@gmail.com}
    }

\begin{document}

\maketitle

\begin{abstract}
Transformers are state-of-the-art deep learning models that are composed of stacked attention and point-wise, fully connected layers designed for handling sequential data. Transformers are not only ubiquitous throughout Natural Language Processing (NLP), but also have recently inspired a new wave of Computer Vision (CV) applications research. In this work, a Vision Transformer (ViT) is finetuned  to predict the state variables of 2-dimensional Ising model simulations. Our experiments show that ViT outperforms state-of-the-art Convolutional Neural Networks (CNN) when using a small number of microstate images from the Ising model corresponding to various boundary conditions and temperatures. This work explores the possible of applications of ViT to other simulations and introduces interesting research directions on how attention maps can learn the underlying physics governing different phenomena. 
\end{abstract}

\section{Introduction}~\label{intro}
The Ising  model is regarded as the simplest theoretical framework to describe ferromagnetism, and understand phase transitions~\cite{Ising1925}. The 2D Ising model is a mathematical model of atomic spins on a lattice which exhibits a phase transition that can be computed analytically~\cite{onsanger2d}. The system undergoes a second order phase transition at the critical temperature $T_c$. In particular, when $T<<T_c$, it undergoes spontaneous magnetization, and this phenomena characterizes ferromagnetism or the ordered state. The local interaction between spins is ultimately responsible for this behavior. For high temperatures, $T >> T_c$  the system is in the disordered or the paramagnetic state. In this case, there are no long-range correlations between the spins. Ferromagnetism is a collective phenomena which occurs when the spins of atoms in a lattice align such that the associated magnetic moments all point in the same direction.
Analytical and numerical solutions for the Ising model have been extremely important in the understanding of phase transitions and ferromagnetism. However, in many cases, these solutions are extremely difficult to formulate and compute. Thus various machine learning (ML) methods have been used to understand these phase transitions. 

Challenges in applying ML techniques to dynamical Ising models include the large number of interacting degrees of freedom and the so-called quenched disorder~\cite{Radzihovsky}, the latter of which might result in a system evolving without temporal significance. Attempts to overcome these problems in Ising like systems using ML date back to~\cite{LITTLE1974, Hopfield1982, Gardner_1988}. Recently, the successes of deep learning and its applications to complex physical systems have reinvigorated interest in the field~\cite{Shiina_2020,dedieu2021sampleefficient, Azizi2021, Morningstar, Carrasquilla2017, Sprague2021}. All of the recent work has been focused on using CNN for understanding the Ising systems near the critical temperature~\cite{Fukushima_2021, Efthymiou_2019, Tanaka_2017, Alexandrou2020, PhysRevB.103.134422, PhysRevE.102.053306}. Several authors have also used CNNs in Generative Adversarial Networks (GAN)~\cite{liu2017simulating} and in Variational Autoencoders~\cite{Walker2020, PhysRevResearch.2.023266} to generate images that simulate the system near the critical temperature. Given the success of transformers in CV, it is natural to ask whether a ViT can match or outperform a CNN in understanding the phase transitions in an Ising model.  

In this work, our contributions are the following: \textbf{(a)} we created a custom-made suite to generate high resolution Ising grid images for a large number of systems with different boundary conditions at various temperatures, \textbf{(b)} a ViT is fine-tuned on these images to predict the state variables corresponding to each of the simulation's experimental constraints, and finally, \textbf{(c)} to the best of our knowledge, this is the first time ViT has been used to understand classical statistical mechanical phenomenon in the Ising model and we show that ViT outperforms CNN based architectures like ResNet-18 and ResNet-50 when using a small number of labeled images. 

\section{Ising Model}~\label{ising}
\label{sec:headings}
The Hamiltonian of a system expresses the total energy of the system in question. Classically, the Hamiltonian is understood as the sum of the kinetic and potential energies. In the case of the two-state Ising Model, the standard  Hamiltonian, $\mathcal{H}(\sigma)$, reads
\begin{equation}\label{eq:1}
\mathcal{H}(\mathbf{\sigma}) = -\sum_{\langle i,j\rangle} J_{i,j} \sigma_i \sigma_j - B\sum_{i} \sigma_i 
\end{equation}
where $J_{i,j}$ is the interaction strength between the $i$th and $j$th spin sites, $B$ is the external magnetic field, and $\sigma_i = \pm 1 $ is the $i$th spin on the grid usually taken, as is the case here, to be an $m \times n$ lattice. The sum ${\langle i,j\rangle}$ is taken over all sites without double counting, and typically, the interaction $J_{i,j}$ is constant for all nearest neighbors across the lattice. 
In all our  systems, $\mathcal{H}(\sigma ) $ is defined according to the availability of spin sites in a restricted finite volume. Starting at T$_c$ for Ising Models, the order parameter of the second-order phase transitions increases continuously from zero. 
 Second-order phase transitions are characterized by a high temperature phase with an average zero magnetization (disordered phase) and a low temperature phase with a non-zero average magnetization. This is very well demonstrated in the Metropolis-Hastings type Ising model when observing the continuous increase of the magnetization at a ferromagnetic-paramagnetic phase transition. When J is positive, the spins align parallel to one another, indicating ferromagnetic behavior. On the other hand, when coupling constant J is negative, nearest neighbor spins are in their lowest energy state by aligning themselves anti-parallel. In that case, the system is anti-ferromagnet. The properties of the anti-ferromagnetic 2D Ising model behaves almost identically to the ferromagnetic Ising model. However, in the lowest energy state $T\approx 0$ a checkerboard-like configuration represents the lowest energy state in contrast to the ferromagnetic case where energy is minimized with all spins in one direction or the other (fig~\ref{fig:fig33}).

\begin{table}[t]
	\centering
	\begin{tabular}{l  c c c c}
\toprule
	& FSbCR & SbCR & Cr & SpCR \\
	\midrule
	Train & 220 & 80 & 90 & 210 \\
	Validation & 100 & 50 & 60 & 90 \\
	Test & 150 & 50 & 70 & 130 \\
	\bottomrule
	\end{tabular}
	\caption{Data distribution: Number of images in each bin for each boundary condition.}
	\label{tab:data_dist}
	\end{table}
 \section{Datasets}
 In this section we will describe the datasets used in our experiments. For this work, we developed a custom suite package to create a diversified set of Ising grid images. These images were generated using the single-flip Metropolis-Hastings algorithm with random temperature variations (between 0K-4K) and three different boundary conditions: \textbf{(a)} Periodic boundary conditions (ferromagnetic and anti-ferromagnetic systems): Converts the plane square lattice system into a torus lattice, \textbf{(b)} Anti-periodic boundary conditions (both directions): The sign of the coupling constant is reversed at the top/bottom and left/right boundaries of the lattice grid, and \textbf{(c)} Skewed $\pm$ boundary conditions: Imposes periodic boundary conditions on a lattice except for at the top and bottom rows.

\begin{figure}[h!]
    \centering
	\includegraphics[width=.7\linewidth]{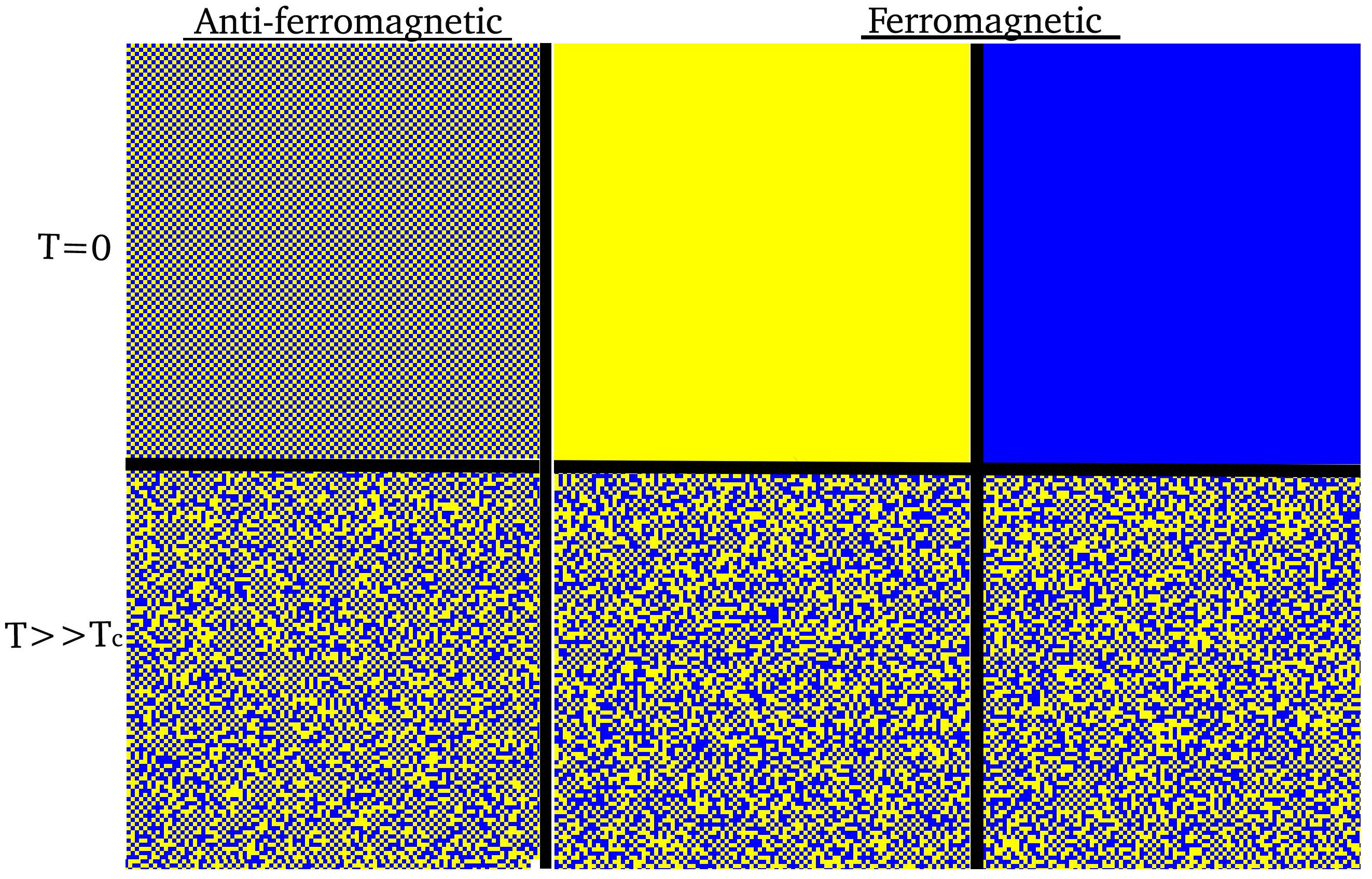}
	\caption{Images of ferromagnetic and anti-ferromagnetic at T = 0 (distinct ground state energy) and T = 4 (disordered system) }
	\label{fig:fig33}
\end{figure}

The experiments for the Ising model are carried out on a $100$x$100$ two dimensional lattice. Each lattice point allows for either spin up or spin down, hence there exist $2^{100\text{x}100}$ potential configurations. We ensure equilibrium, or thermalization is reached between temperature steps by only collecting data after $500$ Monte Carlo sweeps. So we wait $750$ (cushion for equilibrium) Monte Carlo sweeps prior to the subsequent calculation~\cite{landau_binder_2014}. A strong proxy for determining whether or not equilibrium has been reached at a new temperature is elucidated by plotting the average magnetization per spin after each implementation of the Metropolis algorithm against the number of iterations. 

\begin{figure}[h!]
	\centering
	\includegraphics[width=\textwidth]{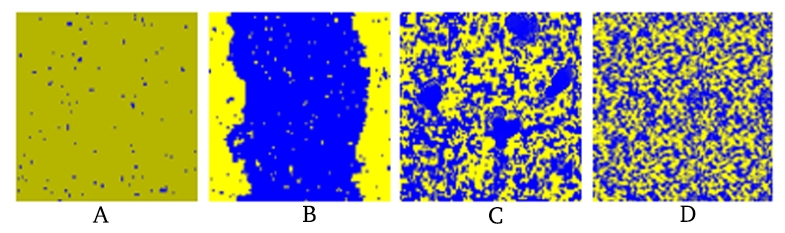}
	\caption{Various Normalized Images Representing Micro-states. The blue area correspond to "spin up" region and the yellow area correspond to "spin down" region. Fig A, B, C, D belongs to FSbCR, SbCR, Cr, SpCR classes respectively.}
	\label{fig:ising_fig}
	\end{figure}

 We generated $1300$ images for each boundary condition with  step size increments of 0.01K. These images were classified into $4$ bins of dissimilar sizes. It is important to point out that for temperatures far above the critical temperature ($T >> 2.27$K), the models tend toward purely noisy systems. The unequal bin sizes were selected to show the effectiveness of the classifier in the small intervals around the critical temperature, and to demonstrate the reliability of  predictions in qualitatively differentiable subregions: 0K-1.055K which we call the \textbf{far sub-critical region} (FSbCR); 1.055K-2.119K deemed the \textbf{sub-critical region} (SbCR); 2.119K-2.320K as the \textbf{critical region} (CR); and 2.320K-4.0K as the \textbf{super-critical region} (SpCR). Bins were selected to show the ability/sensitivity of the model to identify images in a small neighborhood around the critical temperature (2.27K). This sensitivity is demonstrated in the fact that the size of critical region bin is deliberately selected to be at most 30\% the size of the other bins. We are interested in developing a ML model that can accurately predict the label for the image since we believe that such a system can potentially identify characteristic phenomena associated with known critical regions. Interpreting such a model can shed light on what features are associated with multiple critical regions, thus allowing us to look for similar ones in other systems. An example of our dataset is in Figure~\ref{fig:fig33} and~\ref{fig:ising_fig}, whereas the data distribution for various bins are in Table~\ref{tab:data_dist}. For more figures, please see Appendix. 

\section{Methods}~\label{model}
In this section we will describe our method. We start by giving a short introduction to the Vision Transformer and describe how we train the ViT model for our downstream classification task. 
\subsection{Vision Transformers}~\label{vit}
 Transformers have become the de-facto sequence-to-sequence architecture in NLP after their introduction in the landmark paper~\cite{Vaswani2017}. However, attention-based mechanisms in CV have only been used in conjunction with CNNs. CNNs have been used extensively in CV, as they enjoy inductive biases like translation invariance and a locally restricted receptive field. However, Dosovitskiy et al.~\cite{Dosovitskiy2020} successfully introduced Vision Transformer, a transformer-based architecture, to CV which outperforms state-of-the-art CNNs on various downstream CV tasks. Transformers lack the inductive biases of CNNs, and by design they cannot process grid-structured data. The authors cleverly convert a spatial, non-sequential signal by splitting an image into a sequence of flattened, non-overlapping patches. The success of ViT in CV has led to an explosion of research in creating parameter efficient transformers~\cite{CVT}, and faster training of self-supervised transformers~\cite{DINO, AndreasSteiner}.

In our work, we used the ViT-Base model~\cite{Dosovitskiy2020} from the Hugging Face library~\cite{wolf2020huggingfaces}. We used PyTorch and the Hugging Face Library to train our models on a NVIDIA V100 16GB GPU.

\subsection{Finetuning Vision Transformers}~\label{ft_vit}
We started with the checkpoint of ViT which was pre-trained on ImageNet-21k (a collection of 14 million images and 21k classes) and further fine-tuned on ImageNet (a collection of 1.3 million images and 1,000 classes). We then fine-tuned this ViT model on our training set for $10$ epochs with cross-entropy loss. We used AdamW optimizer~\cite{loshchilov2019decoupled} with learning rate 2e-5 and a weight decay of 5e-6, and early stopping on the validation loss to prevent overfitting. To benchmark our performance, we also fine-tuned a ResNet-18 and ResNet-50 on our data. For a fair comparison, both the ResNets were pretrained on ImageNet-21k and then fine-tuned on ImageNet.

\section{Results}~\label{results}
In this section, we will describe the performance of our model. Table~\ref{tab:results} shows that the ViT outperforms both the ResNets when fine-tuned on such a small dataset. This is an active work in progress as we benchmark ViT on other Ising model simulations like the Blume-Capel model and the q-Potts model. 

\begin{table}[h!]

	\centering
	\begin{tabular}{l  c c c}
	\toprule
System/Boundary Conditions & ViT & ResNet-18 & ResNet-50 \\
		\midrule
Periodic (Ferromagnetic) & $\mathbf{0.934 \pm .008}$ & $0.865 \pm .034 $& $0.907 \pm .021$ \\

Periodic (Anti-ferromagnetic)	 &  $\mathbf{0.935 \pm .012}$ & $0.906 \pm .016$  & $0.899 \pm .026$ \\

Skewed (Ferromagnetic) 	 & $\mathbf{0.931 \pm .021}$ & $.886 \pm .019$ & $0.917 \pm .009$\\

Anti-Periodic (Ferromagnetic) & $\mathbf{0.921 \pm .013}$ & $0.91 \pm .021$ & $0.917 \pm .008$ \\
	\bottomrule
	\end{tabular}
		\caption{Table showing the Macro F1 scores (average and standard deviation over 5 trials) of various models on our test sets}
	\label{tab:results}
\end{table}

\section{Conclusion}~\label{conclusion}
We developed a custom suite to generate images from various Ising model simulations at different temperatures and with different boundary conditions. We then used these images to fine-tune a pre-trained ViT and demonstrated that ViT can outperform the current state-of-the-art CNNs. This is still a work in progress as we extend our work to other simulations and applications; to systems undergoing topological phase transition, e.g. XY model, systems widely used to model behaviors of complex systems, e.g. q-Potts model, and the transverse field quantum Ising model (continuous imaginary-time). One of our main motivations in using ViT is to use the attention maps for interpretability. Our goal is to use the attention maps to understand Ising model phase transitions from a visual pattern perspective by looking at Ising configurations, with the different boundary conditions generating qualitatively different looking configurational patterns. We believe this will raise interesting research questions as we try to relate conventional physical approaches usually applied to the Ising problem with the new ideas in computer vision.

\section{Broader Impact}~\label{impact}
The past year has produced a unprecedentedly rapid and cross-disciplinary increase in the study and applications of CV systems across almost every corner of life and industry. Much of the interest is due to the extensive and seemingly endless applicability of transformers and other state of the art deep learning frameworks. In particular, our work shows that state-of-the-art research in machine learning can be used in understanding fundamental sciences and  potentially help in discovering new natural phenomena. However we must be mindful of harmful effects of using these computer vision technologies in society. There are well known examples where these computer vision technologies has mischaracterized people of color and sometimes it has led to severe consequences like harassment towards Black and Brown people. Situations like this can be mitigated with careful focus at each step of the development process. Moreover more research needs to be done in interpreting these models and at the end of the day we should acknowledge that our methods in understanding the predictions of these black box models are incomplete. Thus we must be mindful of how we use technology when we wish to introduce new, powerful, and exciting technologies into society.

\begin{ack}
We would like to thank Christina Tang (contact: christina.f.tang@gmail.com) for technical assistance and for refactoring our Ising model simulations suite code into a fully customizable experimentation suite. We would like to thank Tamiko Jenkins (contact: tommymjenkins@gmail.com) for her tireless efforts and input while proof--reading and editing this manuscript. We would also like to thank John Weeks and Yigit Subasi for helpful comments and suggestions on an earlier version of the manuscript.
\end{ack}


\bibliographystyle{unsrt}
\bibliography{references} 
\newpage









 \appendix

\section{Appendix}

\subsection{Monte Carlo Sampling Algorithm}
 \begin{figure}[h!]
	\centering
	\includegraphics[width=\textwidth]{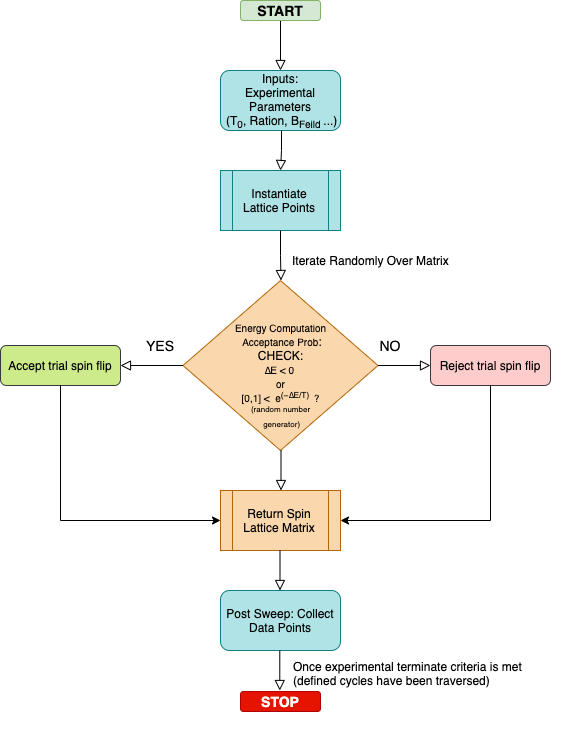}
	\caption{Flow-chart depicting a single sweep though the Metropolis algorithm}
	\label{fig:ising_fig2}
	\end{figure}

\subsection{Normalized Micro-state Image Panels}
 \begin{figure}[h!]
	\centering
	\includegraphics[width=\textwidth]{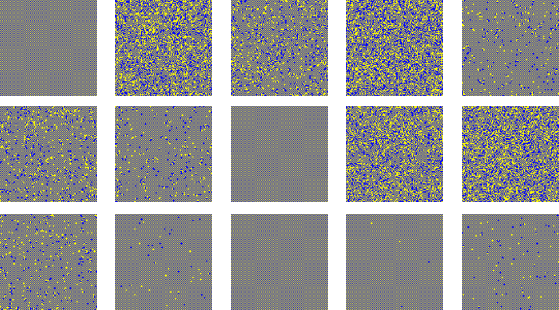}
	\caption{Anti-Ferromagnetic Ising Model With Periodic Boundary Conditions}
	\label{fig:ising_fig2}
	\end{figure}

 \begin{figure}[h!]
	\centering
	\includegraphics[width=\textwidth]{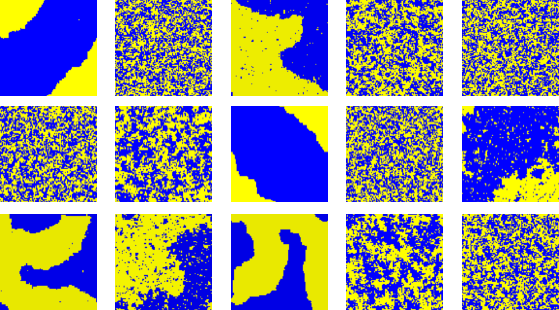}
	\caption{Ising Model With Anti-Periodic Boundary Conditions}
	\label{fig:ising_fig5}
	\end{figure}

 \begin{figure}[h!]
	\centering
	\includegraphics[width=\textwidth]{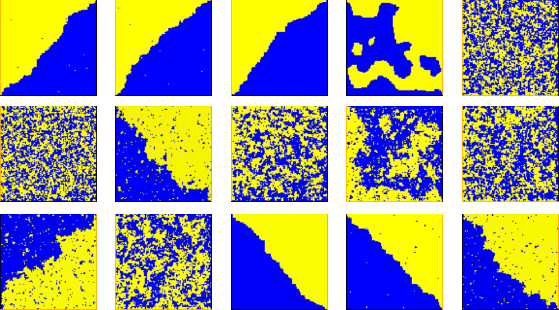}
	\caption{Ising Model With Skewed $\pm$ Boundary Conditions}
	\label{fig:ising_fig6}
	\end{figure}

\clearpage

\end{document}